\documentclass[conference]{IEEEtran}
\IEEEoverridecommandlockouts
\usepackage{cite}
\usepackage{amsmath,amssymb,amsfonts}
\usepackage{graphicx}
\usepackage{textcomp}
\usepackage{xcolor}
\usepackage{comment}
\usepackage{algorithm}
\usepackage{algpseudocode}
\usepackage{subfigure}
\usepackage{tabularx}
\usepackage{booktabs}
\usepackage{array}
\usepackage{comment}
\def\BibTeX{{\rm B\kern-.05em{\sc i\kern-.025em b}\kern-.08em
    T\kern-.1667em\lower.7ex\hbox{E}\kern-.125emX}}
\begin{document}

\title{In-network Attack Detection with Federated Deep Learning
in IoT Networks: Real Implementation and Analysis\\

}
\author{\IEEEauthorblockN{Devashish Chaudhary}
\IEEEauthorblockA{\textit{School of IT} \\
\textit{Deakin University}\\
Geelong, Australia \\
s224281473@deakin.edu.au}
\and
\IEEEauthorblockN{Sutharshan Rajasegarar}
\IEEEauthorblockA{\textit{School of IT} \\
\textit{Deakin University}\\
Geelong, Australia \\
sutharshan.rajasegarar@deakin.edu.au}
\and
\IEEEauthorblockN{Shiva Raj Pokhrel}
\IEEEauthorblockA{\textit{School of IT} \\
\textit{Deakin University}\\
Geelong, Australia \\
shiva.pokhrel@deakin.edu.au}
\and
\IEEEauthorblockN{Lei Pan}
\IEEEauthorblockA{\textit{School of IT} \\
\textit{Deakin University}\\
Geelong, Australia \\
leipanresearch@gmail.com}
\and
\IEEEauthorblockN{Ruby D}
\IEEEauthorblockA{\textit{School of Computer Science and Engineering} \\
\textit{Vellore Institute of Technology}\\
Vellore, India \\
ruby.d@vit.ac.in}
}

\maketitle

\begin{abstract}
The rapid expansion of the Internet of Things (IoT) and its integration with backbone networks have heightened the risk of security breaches. Traditional centralized approaches to anomaly detection, which require transferring large volumes of data to central servers, suffer from privacy, scalability, and latency limitations. This paper proposes a lightweight autoencoder-based anomaly detection framework designed for deployment on resource-constrained edge devices, enabling real-time detection while minimizing data transfer and preserving privacy. Federated learning is employed to train models collaboratively across distributed devices, where local training occurs on edge nodes and only model weights are aggregated at a central server. A real-world IoT testbed using Raspberry Pi sensor nodes was developed to collect normal and attack traffic data. The proposed federated anomaly detection system, implemented and evaluated on the testbed, demonstrates its effectiveness in accurately identifying network attacks. The communication overhead was reduced significantly while achieving comparable performance to the centralized method. 
%The communication overhead was reduced by 91.6\%, achieving comparable performance to the centralized method.
\end{abstract}

\begin{IEEEkeywords}
Internet of Things, network attack detection, anomaly detection, federated learning, transfer learning, edge computing, privacy preservation.
\end{IEEEkeywords}

\section{Introduction}
The widespread adoption of the Internet of Things (IoT) across diverse application domains has led to a substantial increase in the number of interconnected devices and the volume of data generated and exchanged. According to market projections, the global number of IoT connections is expected to reach 38.8 billion by 2029 \cite{Markets}. While this connectivity facilitates automation, monitoring, and intelligent decision-making, it concurrently amplifies the attack surface, thereby increasing the susceptibility of IoT networks to cyber threats. The heterogeneous and distributed nature of IoT infrastructures, combined with the limited computational and storage capabilities of edge devices, presents significant challenges for the deployment of traditional centralized security mechanisms.

Conventional network intrusion detection systems (NIDS) typically rely on transferring raw data from distributed IoT nodes to centralized servers for model training and anomaly detection. This centralized paradigm introduces critical limitations, including increased communication overhead, latency, scalability issues, and heightened risks to data privacy and network integrity. Furthermore, in adversarial settings, the compromise of a single edge node may lead to broader breaches if sensitive data is exposed or leveraged for lateral attacks within the network.

To address these limitations, we propose a federated anomaly detection framework that integrates lightweight deep learning models, specifically autoencoders, directly onto resource-constrained edge devices. Our framework enables real-time detection of anomalous behavior \( \mathcal{B}(t) \) at the edge by minimizing the need for data transmission. Let \( x_t \) represent the input feature vector at time \( t \), and \( \hat{x}_t \) denote the reconstruction of \( x_t \) generated by the autoencoder. The detection of an anomaly is determined based on the reconstruction error \( \mathcal{E}(t) = \| x_t - \hat{x}_t \|^2 \), where values of \( \mathcal{E}(t) \) exceeding a predefined threshold \( \tau \) indicate anomalous behavior:
\[
\mathcal{E}(t) > \tau \Rightarrow \text{Anomaly detected}.
\]
By localizing the model on edge devices, our approach reduces communication overhead and preserves data privacy, as only the model parameters, rather than raw data, are transmitted for global model updates.
 To support scalable and privacy-preserving model updates, we adopt a federated learning (FL) strategy~\cite{9079513}, wherein each edge node trains a local model using its own data and transmits only model parameters (e.g., weights) to a central aggregator. The global model is subsequently updated using the Federated Averaging (FedAvg) algorithm~\cite{mcmahan2017communication}, defined as
\begin{equation}
    W_{t+1}^{\text{global}} = \frac{1}{K} \sum_{k=1}^{K} W_{t}^{(k)},
\end{equation}
where \( W_{t}^{(k)} \) represents the model parameters of the \(k\)-th device at iteration \(t\), and \(K\) is the total number of participating devices. The aggregated global model \( W_{t+1}^{\text{global}} \) is then redistributed to all the devices for subsequent training rounds.

The primary contributions of this work are as follows:

    \begin{itemize}
        \item \textit{Experimental IoT testbed design \& deployment:} We design and implement a hierarchical IoT network using Raspberry Pi devices equipped with XBee modules. Normal network behaviour is recorded under realistic operating conditions, and relevant features are extracted for training and evaluation.

        \item \textit{On-device anomaly detection architecture:} We develop and deploy a resource-efficient autoencoder-based model on edge devices to perform unsupervised anomaly detection in real-time, mitigating the need for centralized data collection.

        \item \textit{Federated Learning (FL) integration:} We adapt the FedAvg algorithm to enable distributed, collaborative anomaly detection model training across IoT devices, ensuring privacy-preserving and communication-efficient operation suitable for the constrained environment.

\item \textit{Attack emulation using Zigbee protocol vulnerabilities:} We implement several redirection attacks by extending and exploiting the attention (AT) command interface of the Zigbee protocol. These attacks are logged and utilized for evaluating detection performance.

        \item   \textit{Transfer learning enhancement:} We incorporate transfer learning to improve the adaptability and generalization of local models across heterogeneous devices and varying traffic conditions.
    \end{itemize}

Our experimental evaluation demonstrates that the proposed federated anomaly detection framework achieves high accuracy in identifying diverse attack scenarios, while preserving user privacy and reducing communication overhead. This work contributes to the development of scalable, privacy-aware security solutions for next-generation IoT systems.

\section{Literature}

Federated learning and anomaly detection have garnered significant attention in the context of IoT security due to the growing need for decentralized, privacy-preserving solutions. Rajasegarar et al. proposed distributed anomaly detection schemes using hyperspherical \cite{rajasegarar2014hyperspherical} and hyperellipsoidal \cite{rajasegarar2014ellipsoidal} cluster-based algorithms to reduce communication overhead while maintaining accuracy. Kanthuru et al. \cite{kanthuru2022cyber} introduced a machine learning-based framework for cyberattack detection in IoT networks using Raspberry Pi nodes and ZigBee modules. While effective, the detection process was performed offline, without in-network or on-device learning capabilities. These methods, however, did not employ deep learning or federated model updates.

Ficco et al. \cite{ficco2024federated} proposed an efficient and scalable learning framework for IoT devices that combines FL and transfer learning (TL), demonstrating improved performance over traditional FL and TensorFlow Lite in classification and regression tasks.

Zeng et al. \cite{zeng2023federated} addressed privacy concerns in FL by introducing a fully homomorphic encryption method based on the Conjugate Search Problem (CSP), enabling secure transmission of model parameters without degrading performance. Korkmaz et al. \cite{korkmaz2022evaluation} evaluated multiple FL strategies, identifying FedAvg as an effective optimization technique for distributed learning. Similarly, Idrissi et al. \cite{idrissi2023fed} developed Fed-ANIDS, which combines autoencoder-based anomaly detection with FL, achieving high accuracy while preserving client data privacy. However, their work lacked real-world deployment on resource-constrained devices.

Piracha et al. \cite{piracha2019insider} investigated AT-command-based attacks using ZigBee sensor nodes but did not explore federated or on-device learning techniques for detection. Ahmed et al. \cite{ahmed2021federated} focused on federated deep learning for heterogeneous edge environments, assigning model complexities based on client resources. Their strategy effectively mitigated straggler effects and optimized training, though it did not include implementation or evaluation in a real IoT testbed. While existing work has made progress in federated and distributed learning for IoT anomaly detection, gaps remain in real-time, on-device implementation and evaluation within realistic network environments. Our work addresses this by deploying a federated, autoencoder-based anomaly detection framework on a real IoT testbed comprising resource-constrained devices.

\section{Proposed Federated Anomaly Detection}

    \begin{figure*}[t]
        \centering
        \subfigure[]{\includegraphics[width=0.32\textwidth]{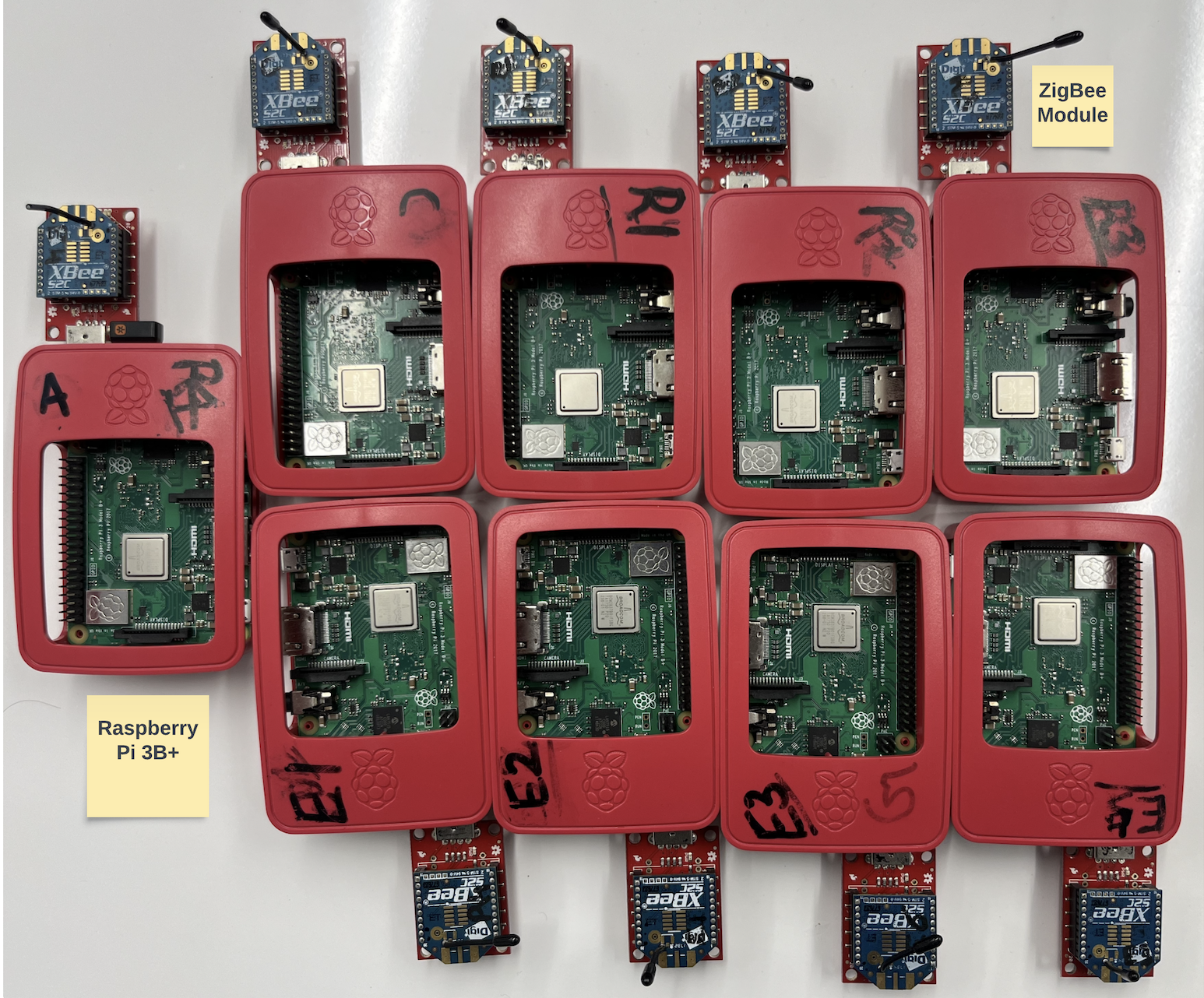}}
        \hfill
        \subfigure[]{\includegraphics[width=0.32\textwidth]{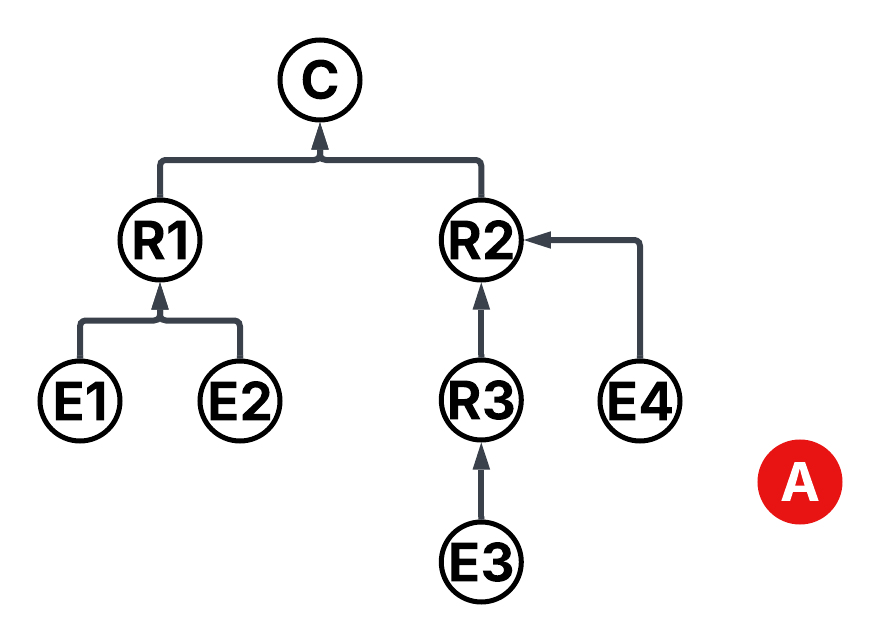}}
        \hfill
        \subfigure[]{\includegraphics[width=0.32\textwidth]{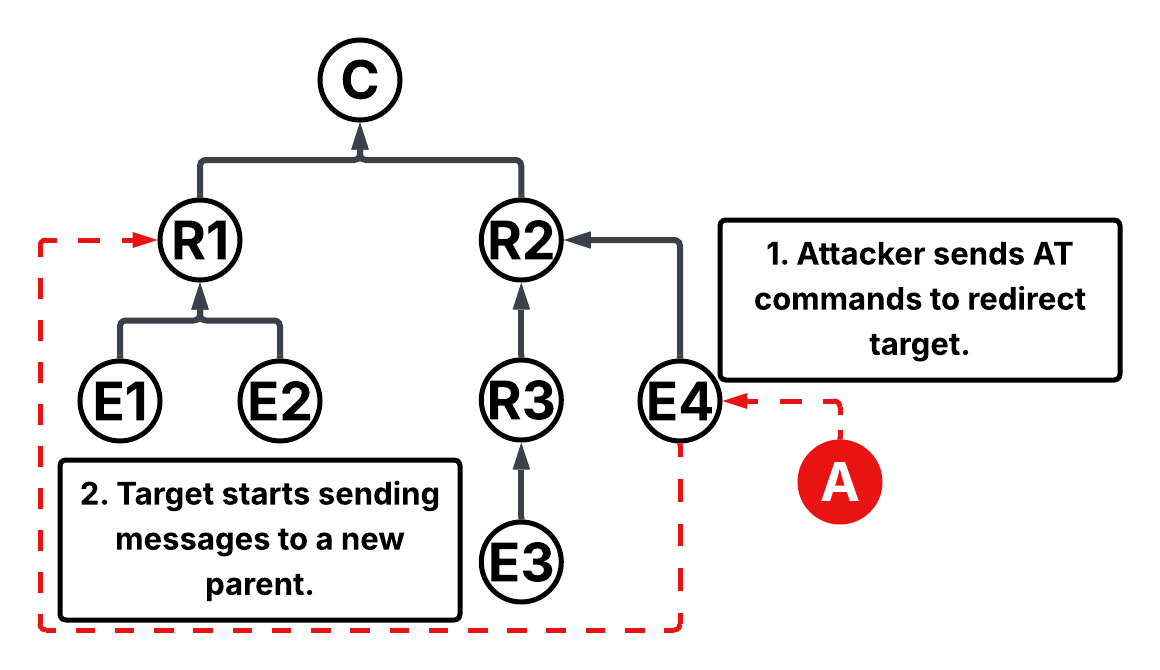}}
        \caption{(a) Raspberry Pi 3B+ with ZigBee (b) Network Topology (c) Redirection Attack.}
        \label{fig:NetworkSensorAndAttack}
    \end{figure*}

    A real-world IoT testbed is designed and constructed using a hierarchical topology composed of multiple resource-constrained sensor nodes. Each node is built on a Raspberry Pi 3B+ platform and equipped with an XBee S2C ZigBee radio module \cite{XBee} to enable wireless communication. One node is designated as an attacker to simulate security threats and facilitate the collection of both benign and malicious traffic data. Initially, network traffic logs are recorded under normal operating conditions, and relevant features are extracted for training a deep learning-based anomaly detection model. Subsequently, various attack scenarios are executed to generate malicious traffic, which is logged and used to evaluate the detection system.

    To improve the efficiency and scalability of model training, FL and TL are employed \cite{zhao2020network}. In the FL framework, each edge device independently trains a local model and transmits only the model parameters to a central coordinator. The coordinator aggregates these updates by adapting the FedAvg algorithm to construct a global model, thus preserving data privacy at the device level. TL is utilized to leverage knowledge from previously trained models, reducing the computational burden on resource-constrained devices and accelerating the training process.

    \textbf{Network Topology:} Our network topology consists of Raspberry Pis, comprising a coordinator (C), three routers (R1, R2 and R3), four edge devices (E1, E2, E3 and E4), and a malicious node (see Fig.~\ref{fig:NetworkSensorAndAttack}(b)). Each Raspberry Pi is equipped with an XBee module for wireless communication (see Fig.~\ref{fig:NetworkSensorAndAttack}(a))  \cite{XBee}. The network configuration can be managed using the XCTU software. The setup and wireless communication via the ZigBee modules are managed by a Python program that leverages the Digi API (Application Programmer Interface) \cite{PythonLibrary}.

    \textbf{Redirection Attacks:} In our experimentation, we implemented redirection attacks wherein the destination address of a target node was altered to point to a new parent node (see Fig.~\ref{fig:NetworkSensorAndAttack}(c)). This manipulation was achieved through the exploitation of AT (attention) commands, defined by the ZigBee protocol \cite{vaccari2017remotely} and an attacker node operating within the same PAN ID. 

    Fig.~\ref{fig:NetworkSensorAndAttack}(c) shows an example of a redirection attack scenario implemented. It can be observed that the edge node (E4), which originally sent messages to router R4, was targeted, and its destination address was altered to router R1 using the AT commands. Consequently, it will commence sending messages to R1 instead of R4.

    \textbf{Data Logging Format:} In our network system, each device keeps a record of the messages it handles, including details such as timestamps, node IDs, and status codes. For example, when an end device sends a packet, it records the XBee ID and the time of transmission. Routers similarly log information such as received data and timestamps, along with sender and destination IDs.
    
    To automate log collection, a process is implemented where each node sends its logs to the coordinator. This ensures that the coordinator receives comprehensive information about the packet's journey through the network. Edge devices are programmed to send packets every one second. A typical log entry at each device will have the following format:

    \begin{itemize}
        \item Coordinator:

        \texttt{E3>R3,2024-04-26 13:36:10.273312,
        2024-04-26 13:36:10.336880,
         R3>R2,2024-04-26 13:36:10.369257,
         2024-04-26 13:36:10.488817,
         R2>C,2024-04-26 13:36:10.522766,
         2024-04-26 13:36:10.787851}

        This entry demonstrates that the packet has originated from end device E3 at 
        \texttt{2024-04-26 13:36:10.273312}, received at router R3 at 
        \texttt{2024-04-26 13:36:10.336880}, sent from router R3 at 
        \texttt{2024-04-26 13:36:10.369257}, then received at router R2 at 
        \texttt{2024-04-26 13:36:10.488817}, sent from router R2 at 
        \texttt{2024-04-26 13:36:10.522766} and ultimately reached coordinator C at 
        \texttt{2024-04-26 13:36:10.787851}.
        This approach streamlines log collection for analysis and troubleshooting purposes. The logging process at the central (coordinator) node is not required in practice, as it can be accessed from individual nodes separately. Here, it is used for convenience and easy log collection purposes.
         \\

         \item Router:
         
            \texttt{E3>R3,2024-04-26 13:36:10.273312,
            2024-04-26 13:36:10.336880,
            R3>R2,2024-04-26 13:36:10.369257, S:0}

        In this router log entry, the communication flow between devices is detailed. It begins with end device E3 transmitting a message to router R3 at \texttt{2024-04-26 13:36:10.273312}. Subsequently, R3 forwards this message to router R2, as indicated by the timestamp \texttt{2024-04-26 13:36:10.336880}. The entry concludes by noting that this transmission from R3 to R2 was successful, denoted by "S:0" at \texttt{2024-04-26 13:36:10.369257}. This logging method efficiently documents the routing path and success status of each transmission, aiding in log analysis, network troubleshooting, and performance evaluation.
        \\
        \item Edge Device:
        
        \texttt{E3 > R3, 2024-04-26 13:36:10.273312, S:0}

        The entry signifies that edge device E3 successfully sent a message to router R3 at \texttt{2024-04-26 13:36:10.273312}, denoted by "S:0". This logging format succinctly captures the communication event, indicating the source, destination, timestamp, and success status, facilitating efficient monitoring and analysis of device interactions within the network.
    \end{itemize}

    \textbf{Autoencoder:} A neural network used for unsupervised learning, comprising an encoder that compresses input data and a decoder that reconstructs the original input. The goal is to learn a compact representation by minimizing reconstruction errors. In this study, the autoencoder architecture consists of an input layer with 31 features, followed by an encoder with two fully connected dense layers of 32 and 16 neurons, respectively, both using ReLU activation functions. The decoder mirrors this structure with two dense layers of 32 and 31 neurons, using sigmoid activation functions to reconstruct the original input. The model is instantiated using the Keras Functional API, compiled with the Adam optimizer and Mean Squared Error (MSE) loss function. Training is conducted for 100 epochs with a batch size of 32, using a learning rate of 0.001.\\
    
    \textbf{Maths of Autoencoder Formulation}.
Let \( \mathcal{X} \subseteq \mathbb{R}^{31} \) denote the input space of IoT network traffic features. An autoencoder is a neural network comprising two parametrized functions:
\begin{itemize}
    \item \textit{Encoder}: A mapping \( f_\theta: \mathcal{X} \rightarrow \mathcal{Z} \) to latent space \( \mathcal{Z} \subseteq \mathbb{R}^{18} \):
    \[
    \mathbf{z} = f_\theta(\mathbf{x}) = \text{ReLU}\left(\mathbf{W}_e \mathbf{x} + \mathbf{b}_e\right),
    \]
    where \( \theta = \{\mathbf{W}_e \in \mathbb{R}^{18 \times 31}, \mathbf{b}_e \in \mathbb{R}^{18}\} \).

    \item \textit{Decoder}: A mapping \( g_\phi: \mathcal{Z} \rightarrow \mathcal{X} \) for reconstruction:
    \[
    \hat{\mathbf{x}} = g_\phi(\mathbf{z}) = \sigma\left(\mathbf{W}_d \mathbf{z} + \mathbf{b}_d\right),
    \]
    where \( \phi = \{\mathbf{W}_d \in \mathbb{R}^{31 \times 18}, \mathbf{b}_d \in \mathbb{R}^{31}\} \), and \( \sigma(\cdot) \) is the sigmoid activation function.
\end{itemize}
Our design minimizes the \textit{reconstruction loss} over batch \( \mathcal{B} \):
\begin{equation}
    \mathcal{L}(\theta, \phi) = \frac{1}{|\mathcal{B}|} \sum_{\mathbf{x} \in \mathcal{B}} \left\| \mathbf{x} - g_\phi(f_\theta(\mathbf{x})) \right\|^2_2,
    \label{eq:recon_loss}
\end{equation}
optimized via Adam with learning rate \( \eta = 0.001 \), batch size \( |\mathcal{B}| = 32 \), and \( T = 100 \) epochs.

    \textbf{Federated Learning:} We leverage FL by adapting the FedAvg algorithm to train machine learning models across decentralized devices, ensuring data privacy. The FL process involves iterative model training on local data at client devices, followed by the aggregation of model updates at a central coordinator.
As outlined in Algorithm \ref{alg: fed}, at the heart of our solution is aggregating model updates from individual devices, facilitating the convergence of a global model that reflects the collective knowledge from diverse datasets. This iterative process, detailed in Algorithm 1, highlights the core principle of federated learning: enabling collaborative model training without compromising data confidentiality.

\textbf{Maths of the FL Framework}.
Consider \( K \) distributed devices with local datasets \( \{\mathcal{D}_k\}_{k=1}^K \). Let \( \mathbf{w}^{(t)} = \{\theta^{(t)}, \phi^{(t)}\} \) denote global parameters at communication round \( t \). The FedAvg protocol proceeds as:

\begin{itemize}
    \item \textit{Local Training}: Each client \( k \) initializes \( \mathbf{w}_k^{(t)} \leftarrow \mathbf{w}^{(t)} \) and performs stochastic gradient descent (SGD):
    \[
    \mathbf{w}_k^{(t+1)} = \mathbf{w}_k^{(t)} - \eta \nabla_{\mathbf{w}} \mathcal{L}_k(\mathbf{w}),
    \]
    where \( \mathcal{L}_k(\mathbf{w}) = \frac{1}{|\mathcal{D}_k|} \sum_{\mathbf{x} \in \mathcal{D}_k} \left\| \mathbf{x} - g_\phi(f_\theta(\mathbf{x})) \right\|^2_2 \).

    \item \textit{Model Aggregation}: The coordinator computes:
    \begin{equation}
        \mathbf{w}^{(t+1)} = \frac{1}{K} \sum_{k=1}^K \mathbf{w}_k^{(t+1)}.
        \label{eq:fedavg}
    \end{equation}
\end{itemize}
    \begin{algorithm}
    \caption{Adapted Federated Learning Process}
    \label{alg: fed}
    \begin{algorithmic}[1]
    %\caption{Federated Learning Process}
    %\label{alg: fed}
    \State \textbf{Initialization:} Coordinator (C) prepares initial model $W_0$
    \For{each Router $R_i$}
        \If{$R_i$ is a leaf router}
            \State $R_i$ requests and receives $W_0$ from C
        \Else
            \State $R_i$ requests and receives $W_0$ from parent router
        \EndIf
        \State $R_i$ initializes local model with $W_0$
    \EndFor
    \Repeat
        \For{each Router $R_i$ \textbf{in parallel}}
            \State $R_i$ checks for new local data $D_i$
            \If{$D_i$ available}
                \State $R_i$ trains local model $W_i$ on $D_i$
            \EndIf
        \EndFor
        \For{each Router $R_i$}
            \If{training completed}
                \State $sum\_weights \leftarrow W_i$, $count \leftarrow 1$
                \If{$R_i$ has child routers}
                    \For{each child $R_{child}$}
                        \State $R_{child}$ sends weights $W_{child}$ to $R_i$
                        \State $sum\_weights \leftarrow sum\_weights + W_{child}$
                        \State $count \leftarrow count + 1$
                    \EndFor
                \EndIf
                \State $R_i$ sends $sum\_weights$ and $count$ to C
            \EndIf
        \EndFor
        \If{Coordinator received all weights}
            \State C aggregates $sum\_weights$ using $count$ \\ 
              to obtain global model $W_{global}$
            \State C sends completion message to all Routers
        \EndIf
        \For{each Router $R_i$}
            \If{aggregation message received}
                \If{$R_i$ is not a leaf router}
                    \State $R_i$ requests $W_{global}$ from C, updates local model
                    \For{each child $R_{child}$}
                        \State $R_i$ sends $W_{global}$ to $R_{child}$, $R_{child}$ updates model
                    \EndFor
                \Else
                    \State $R_i$ requests $W_{global}$ from parent router, updates model
                \EndIf
            \EndIf
        \EndFor
    \Until{no new data for training}
    \end{algorithmic}
    \end{algorithm}

    \textbf{Transfer Learning:} In this framework, transfer learning enables each client to initialize its local model using a pre-trained model \( \mathbf{w}_0 \), obtained from a related task or domain. Given a local dataset \( \mathcal{D}_k \), client \( k \) fine-tunes the model via gradient descent:
\[
\mathbf{w}_k^{(t+1)} = \mathbf{w}_0 - \eta \nabla_{\mathbf{w}} \mathcal{L}_k(\mathbf{w}_0),
\]
where \( \mathcal{L}_k(\mathbf{w}) \) is the local loss function. This approach incorporates prior knowledge, accelerating convergence and improving generalization in the FL process.

\subsection{Normal and Attack Data Collection}
    Our dataset generation process involved two distinct steps tailored for federated and centralized learning approaches. For federated learning, we computed features at the local level, focusing on routers within our network setup. In contrast, for centralized learning, feature calculations were centralized at the coordinator node, where all the routers in the network send the data packets. For both approaches, features were extracted over one-minute time windows, capturing detailed traffic characteristics. These features include:
\begin{itemize}

    \item \textbf{Delay-based metrics:}
    \begin{itemize}
        \item \textbf{Mean delay:} The average time taken for packets to travel from source to destination within the time window. High mean delay can indicate congestion or anomalies.
        \item \textbf{First hop delay:} The time taken for a packet to reach the first router/hop. It helps detect local network latency issues.
        \item \textbf{Quartiles (Q1, Q2, Q3):} Represent the distribution of delays, capturing variability and skewness in packet travel times. This is useful to detect unusual spikes or patterns in network latency.
    \end{itemize}

    \item \textbf{Shannon entropy:} Calculated over delay and first hop delay to measure uncertainty or variability in network timing. Higher entropy indicates more irregular traffic patterns, which can be indicative of anomalies or attacks.

    \item \textbf{Communication counts:}
    \begin{itemize}
        \item \textbf{Per communication type:} Number of packets exchanged for each communication pair (e.g., device-to-device or device-to-router).
        \item \textbf{Overall count:} Total number of communications in the time window. Sudden spikes or drops can highlight abnormal network activity.
    \end{itemize}

    \item \textbf{Network structure metrics:} 
    \begin{itemize}
        \item \textbf{Average number of hops per communication:} Captures how far packets travel within the network. Changes in average hops can reflect routing anomalies or network attacks that alter typical paths.
    \end{itemize}

\end{itemize}

    These features were vital components of our pre-processing phase, with delays quantified in milliseconds.

    At the coordinator level, there were a total of 31 features extracted. However, for the features at the local level (routers), we utilized a subset of the coordinator-level features, as some of the features were not relevant to edge or router devices. To ensure compatibility with the input dimension required by our deep learning model for federated learning, any missing features at the router level were handled by replacing them with zeros during the pre-processing phase. 

    The features were further normalized using the MinMax scaler to ensure that all values fall within a uniform range and to improve model convergence.
    
\section{IoT Testbed Design and attack scenarios implementation}
%Experimental Setup}
    Our experimental setup comprises a network of 9 sensor nodes designed to collect real-time data from IoT networks under various attack scenarios. Each sensor node is based on Raspberry Pi 3B+ devices, equipped with Digi XBee S2C Zigbee Radio Frequency modules for wireless communication using the ZigBee protocol \cite{komilov2023application}. The network topology follows a hierarchical structure, strategically distributing sensor nodes to capture diverse network traffic patterns. Specifically, one node serves as the coordinator, three nodes function as routers, four nodes operate as edge devices, and one node acts as an attacker.

    Before the logging process began, we ensured that time synchronization was performed in all the Raspberry Pi devices by connecting them to the internet once. Initially, we distributed a pre-trained model on normal network behavior, trained using 1 hour of data, across the nodes to facilitate federated learning. Subsequently, we logged data over a 5-hour period under normal network behavior without any attacks. This 5-hour dataset was used to extract features and train an autoencoder model. In the federated learning setup, each router node trained the model using its local data, and the resulting weights were then transmitted to the coordinator for aggregation. Upon aggregation, the updated global weights were distributed back to the client nodes. Each client node performed training on the new data generated every ten minutes within the 5-hour timeframe, sending updated weights to the coordinator for aggregation after each training cycle.
    
    Concurrently, we also trained a centralized model using the data stored at the coordinator node. After collecting and training the model with normal behavior data, we conducted redirection attacks to test our model's robustness and efficacy. These redirection attacks were designed to simulate real-world scenarios and assess the network's response under different attack conditions. The attacks are classified into three scenarios (see Table~\ref{tab:AttScenarios}): 
    \begin{itemize}
        \item Scenario I: Changing the destination addresses of edge devices within the network. 
        \item Scenario II: Altering the destination addresses of routers.
        \item Scenario III: Directing edge devices and routers to send their data to the attacker node. 
    \end{itemize}

    The Table \ref{tab:AttScenarios} shows the nodes that communicate under normal (non-attack) conditions, and the change in communication happens when the attack is performed. For example, in Table \ref{tab:AttScenarios}, in the second row and second column, the entry $E1 -> R1$ indicates that under normal (non-attack) conditions, node E1 transfers data to router R1. When an attack (scenario-1) is performed, the node E1 traffic is redirected to R2, that is $(E1->R2)$ (as shown in the third row and second column of the table). In this way, under scenario 1, there are 12 attacks implemented. Similarly, 5 attacks for scenario II and 7 attacks for scenario III are implemented, respectively. 

\begin{table*}[ht]
  \centering
  \caption{Attack scenarios: C – Coordinator, A – Attacker, Rx – Router, Ex – Edge/Leaf device.}
  \resizebox{\textwidth}{!}{%
    \begin{tabular}{r|l|l|l|l|l|l|l|l|l|l|l|l|l|l}
    \hline
    \multicolumn{5}{c|}{\textbf{Scenario I}} & \multicolumn{3}{c|}{\textbf{Scenario II}} & \multicolumn{7}{c}{\textbf{Scenario III}} \\
    \hline
    \multicolumn{1}{l|}{\textbf{Normal}} & E1 $\rightarrow$ R1 & E2 $\rightarrow$ R1 & E3 $\rightarrow$ R3 & E4 $\rightarrow$ R2 & \multicolumn{1}{l|}{R1 $\rightarrow$ C} & \multicolumn{1}{l|}{R2 $\rightarrow$ C} & \multicolumn{1}{l|}{R3 $\rightarrow$ R2} & \multicolumn{1}{l|}{E1 $\rightarrow$ R1} & \multicolumn{1}{l|}{E2 $\rightarrow$ R1} & \multicolumn{1}{l|}{E3 $\rightarrow$ R3} & \multicolumn{1}{l|}{E4 $\rightarrow$ R2} & \multicolumn{1}{l|}{R1 $\rightarrow$ C} & \multicolumn{1}{l|}{R2 $\rightarrow$ C} & \multicolumn{1}{l}{R3 $\rightarrow$ C} \\
    \hline
    \textbf{Attack} & E1 $\rightarrow$ R2 & E2 $\rightarrow$ R2 & E3 $\rightarrow$ R1 & E4 $\rightarrow$ R1 & \multicolumn{1}{l|}{R1 $\rightarrow$ R2} & \multicolumn{1}{l|}{R2 $\rightarrow$ R1} & \multicolumn{1}{l|}{R3 $\rightarrow$ R1} & \multicolumn{1}{l|}{E1 $\rightarrow$ A} & \multicolumn{1}{l|}{E2 $\rightarrow$ A} & \multicolumn{1}{l|}{E3 $\rightarrow$ A} & \multicolumn{1}{l|}{E4 $\rightarrow$ A} & \multicolumn{1}{l|}{R1 $\rightarrow$ A} & \multicolumn{1}{l|}{R2 $\rightarrow$ A} & \multicolumn{1}{l}{R3 $\rightarrow$ A} \\
    & E1 $\rightarrow$ R3 & E2 $\rightarrow$ R3 & E3 $\rightarrow$ R2 & E4 $\rightarrow$ R3 & \multicolumn{1}{l|}{R1 $\rightarrow$ R3} & & \multicolumn{1}{l|}{R3 $\rightarrow$ C} & & & & & & & \\
    & E1 $\rightarrow$ C & E2 $\rightarrow$ C & E3 $\rightarrow$ C & E4 $\rightarrow$ C & & & & & & & & & & \\
    \hline
    \end{tabular}%
  }
  \label{tab:AttScenarios}%
\end{table*}

    For each attack scenario, we logged data comprising 20 minutes of normal network behavior, followed by a 5-minute attack period, and then resumed logging another 10 minutes of normal behavior. This systematic approach enabled us to evaluate our model's performance and its ability to detect and mitigate redirection attacks. Also, these scenarios cover all the possible redirection attacks that can occur in the network topology that we have implemented.

    Redirection attacks were chosen because they effectively alter network routing patterns, delays, and communication distributions, which are precisely the behaviors captured by our extracted features (delay metrics, entropy, communication counts, and average hops). While other attacks, such as DoS, DDoS, packet injection, or spoofing exist, their impact on the feature space is often similar in terms of causing abnormal delays, irregular traffic patterns, or spikes in communication counts. Therefore, focusing on redirection attacks allows us to study the detection of significant anomalous behavior while keeping the dataset controlled and manageable. This approach demonstrates that our models can detect anomalous traffic patterns, and the methodology can be extended to other types of attacks in future work.

    \subsection{Defining Threshold to detect attacks}

    In this work on anomaly detection and security in IoT networks, we employ both centralized and federated models for anomaly detection. This requires threshold values to be chosen to determine the normal and anomalies using the autoencoder model. We define thresholds for local devices (routers) based on the reconstruction loss from the validation dataset, which captures normal behavior data recorded for an hour.

\begin{figure}
        \centering
        \includegraphics[width=1\linewidth]{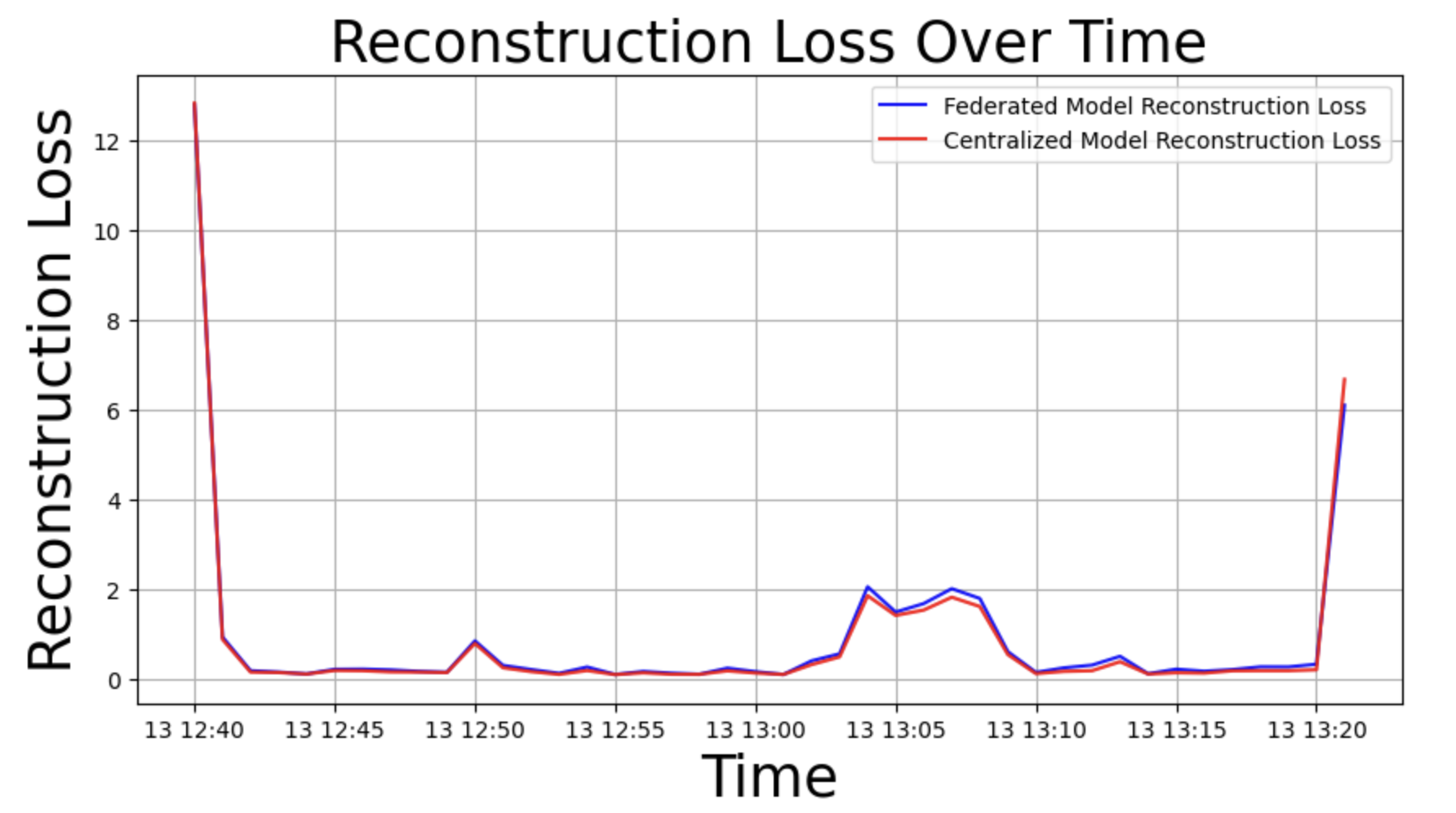}
        \caption{Reconstruction loss/error over time. A small loss is observed during the normal (attack-free) operation, and the loss becomes higher during the attack period (towards the middle).}
        \label{fig:reconstruction_graph}
    \end{figure}

    The graph (see Fig.~\ref{fig:reconstruction_graph}) illustrates the reconstruction loss over time for both the federated and centralized models in your anomaly detection system. The x-axis represents time intervals, and the y-axis shows the reconstruction loss, indicating the model's ability to reconstruct input data, with lower values being better. The blue line represents the federated model, while the red line represents the centralized model. From the given graph, we can see that the federated and centralized models perform very similarly, almost completely overlapping. This graph is for one of the attack scenarios and is shown here as an example to demonstrate its comparable performance. Notably, both models successfully detect the attack, as indicated by the spikes in reconstruction loss, showing the models' sensitivity to anomalies.

    For both the centralized and federated approaches, the threshold for anomaly detection in routers is calculated using the following formula:
    \begin{equation}
        \text{Threshold} = \text{Mean} + k \times \text{Standard Deviation}
    \end{equation}

    Here, the mean and standard deviation are derived from the reconstruction loss of normal behavior data recorded in the validation dataset. The scaling factor 
    $k$ is adjusted to optimize the threshold for anomaly detection across all models and devices.
    
    By leveraging the reconstruction loss from the respective models and tailoring thresholds for each device and the coordinator, we enhance the anomaly detection capabilities of both centralized and federated models while considering the unique characteristics and responsibilities of each component in the IoT network.

\section{Results and Discussion}
    In this section, we present the results of our anomaly detection system using autoencoder models, evaluated through both centralized and federated approaches. The thresholds for anomaly detection were calculated for various values of $k$, ranging from 1 to 4.
    
\subsection{Threshold Calculation}

    For the centralized model, thresholds were determined using data aggregated at the coordinator, providing a global perspective of the network. This comprehensive approach allows for robust anomaly detection but may be less sensitive to localized variations. In contrast, the federated model involved training autoencoders on the data collected by the individual routers themselves (i.e., at source), enabling localized anomaly detection that is more responsive to specific network segments' behavior (i.e., subnet centered with the node). Table~\ref{tab:Thresholds} summarizes the thresholds obtained for both centralized and federated models across different values of $k$.

    \begin{table}[ht]
    \caption{Federated and centralized thresholds}
    \label{tab:Thresholds}
    \centering
    %\scriptsize
    \renewcommand{\arraystretch}{1.1}
    \begin{tabularx}{\columnwidth}{>{\raggedright\arraybackslash}l
                                >{\centering\arraybackslash}X
                                >{\centering\arraybackslash}X
                                >{\centering\arraybackslash}X
                                >{\centering\arraybackslash}X}
    \toprule
    \textbf{Metric} & \textbf{k = 1} & \textbf{k = 2} & \textbf{k = 3} & \textbf{k = 4} \\
    \midrule
    Federated R1   & 0.0342 & 0.0429 & 0.0516 & 0.0603 \\
    Centralized R1 & 0.0014 & 0.0022 & 0.0031 & 0.0039 \\
    \midrule
    Federated R2   & 0.0351 & 0.0450 & 0.0550 & 0.0649 \\
    Centralized R2 & 0.0035 & 0.0058 & 0.0082 & 0.0105 \\
    \midrule
    Federated R3   & 0.0242 & 0.0264 & 0.0286 & 0.0308 \\
    Centralized R3 & 0.0005 & 0.0007 & 0.0009 & 0.0011 \\
    \bottomrule
    \end{tabularx}
\end{table}
\subsection{Performance Metrics}
    In order to evaluate our anomaly detection process for accurately detecting the attacks, as well as to analyze the sensitivity of the model to the values of $k$, we used the accuracy, precision, recall, and F1-score for each threshold value. The formulas used to calculate these metrics are as follows:
    %\begin{equation}
    $\text{Accuracy} = \frac{TP + TN}{TP + TN + FP + FN}$,
    %\end{equation}
    %\begin{equation}
    $\text{Precision} = \frac{TP}{TP + FP}$,
    %\end{equation}
    %\begin{equation}
    $\text{Recall} = \frac{TP}{TP + FN}$,
    %\end{equation}
    %\begin{equation}
    $\text{F1-Score} = \frac{2 \times \text{Precision} \times \text{Recall}}{\text{Precision} + \text{Recall}}$,
    %\end{equation}
    where $TP$ is the number of true positives, $TN$ is the number of true negatives, $FP$ is the number of false positives, and $FN$ is the number of false negatives.

    \begin{table*}[htbp]
    \centering
    %\scriptsize
        \caption{Performance of federated and centralized models (R1, R2, and R3)}
    \label{tab:combined_metrics}
    \resizebox{\textwidth}{!}{%
    \begin{tabular}{ccccccccccccccccc}
    \toprule
    & \multicolumn{4}{c}{\textbf{R1}} & \multicolumn{4}{c}{\textbf{R2}} & \multicolumn{4}{c}{\textbf{R3}} \\
    \cmidrule(lr){2-5} \cmidrule(lr){6-9} \cmidrule(lr){10-13} \cmidrule(lr){14-17}
    \textbf{k}  & \textbf{Acc.} & \textbf{Prec.} & \textbf{Rec.} & \textbf{F1} & \textbf{Acc.} & \textbf{Prec.} & \textbf{Rec.} & \textbf{F1} & \textbf{Acc.} & \textbf{Prec.} & \textbf{Rec.} & \textbf{F1} \\
    \midrule
    \multicolumn{17}{c}{\textbf{Federated method}}\\
    \midrule
    k = 1 &  0.8458 & 0.4440 & 0.9583 & 0.6069 & 0.9333 & 0.7742 & 0.8054 & \textbf{0.7895} & 0.6526 & 0.1809 & 0.8810 & 0.3002 \\
    k = 2 & 0.9369 & 0.6766 & 0.9417 & 0.7875 & 0.9354 & 0.8222 & 0.7450 & 0.7817 & 0.8097 & 0.2908 & 0.8690 & \textbf{0.4358} \\
    k = 3  & 0.9669 & 0.8284 & 0.9250 & 0.8740 & 0.9292 & 0.8649 & 0.6443 & 0.7385 & 0.8127 & 0.2910 & 0.8452 & 0.4329 \\
    k = 4 & 0.9741 & 0.8926 & 0.9000 & \textbf{0.8963} & 0.9250 & 0.8738 & 0.6040 & 0.7143 & 0.8167 & 0.2941 & 0.8333 & 0.4348 \\
    \midrule
    \multicolumn{17}{c}{\textbf{Centralized method}}\\
    \midrule
    k = 1 & 0.9048 & 0.5673 & 0.9833 & 0.7195 & 0.8000 & 0.4343 & 0.9530 & 0.5966 & 0.6123 & 0.1777 & 0.9881 & 0.3013 \\
    k = 2 & 0.9513 & 0.7325 & 0.9583 & 0.8303 & 0.8573 & 0.5227 & 0.9262 & 0.6683 & 0.7432 & 0.2463 & 0.9881 & 0.3943 \\
    k = 3 & 0.9689 & 0.8261 & 0.9500 & 0.8837 & 0.8854 & 0.5867 & 0.8859 & 0.7059 & 0.8308 & 0.3306 & 0.9762 & 0.4940 \\
    k = 4 & 0.9741 & 0.8571 & 0.9500 & \textbf{0.9012} & 0.9052 & 0.6450 & 0.8658 & \textbf{0.7393} & 0.8852 & 0.4219 & 0.9643 & \textbf{0.5870} \\
    \bottomrule
    \end{tabular}%
    }

\end{table*}

    \subsection{Analysis of attack detection performance}
    
    In our evaluation, we compared the performance of centralized and federated learning approaches for anomaly detection in an IoT network. 
    
\begin{figure}[t]
        \centering
        \subfigure[]{\includegraphics[width=1.00\linewidth]{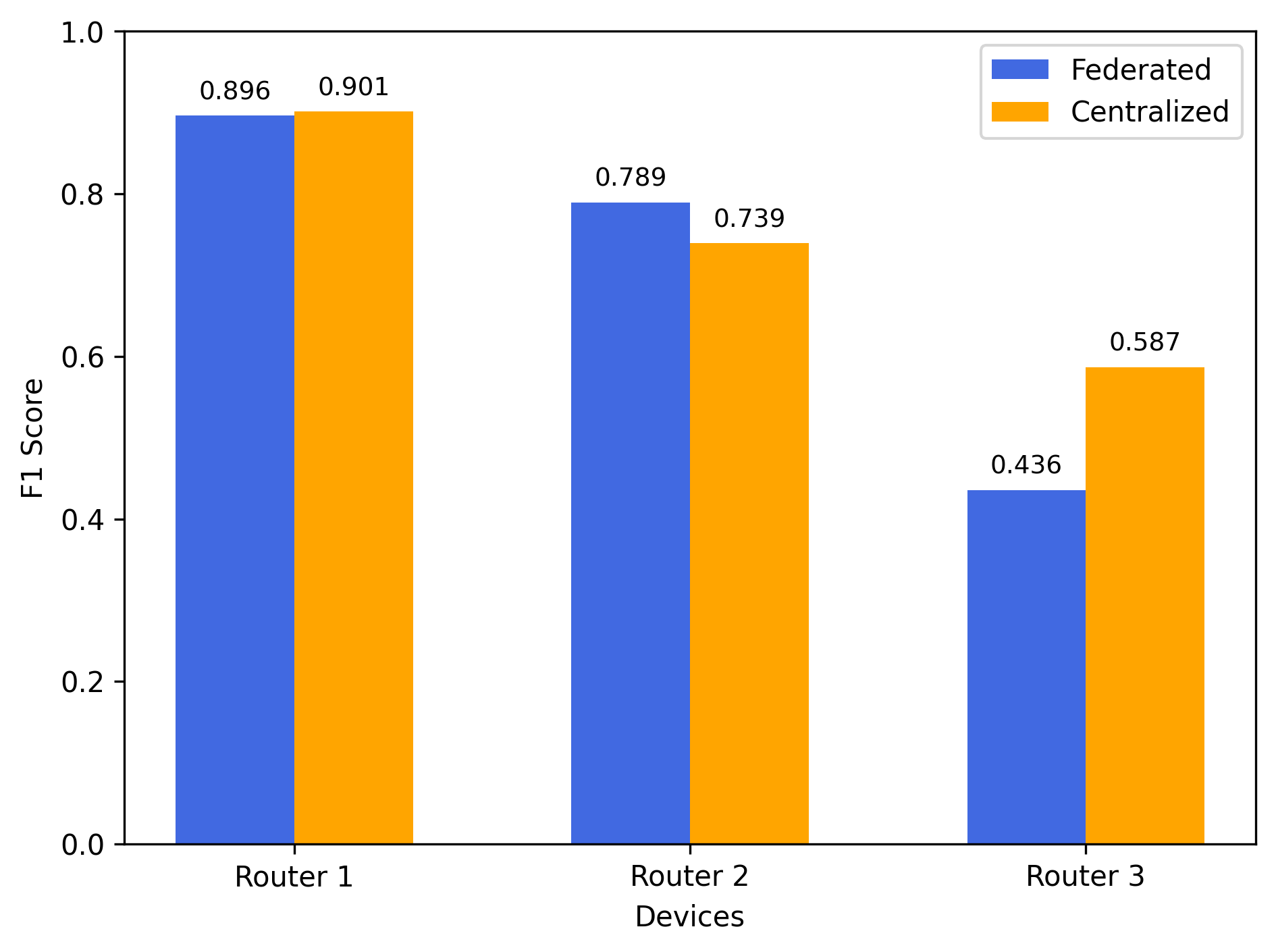}}
        %\label{fig:raspberry}}
        %\hfill
        \subfigure[]{\includegraphics[width=1.00\linewidth]{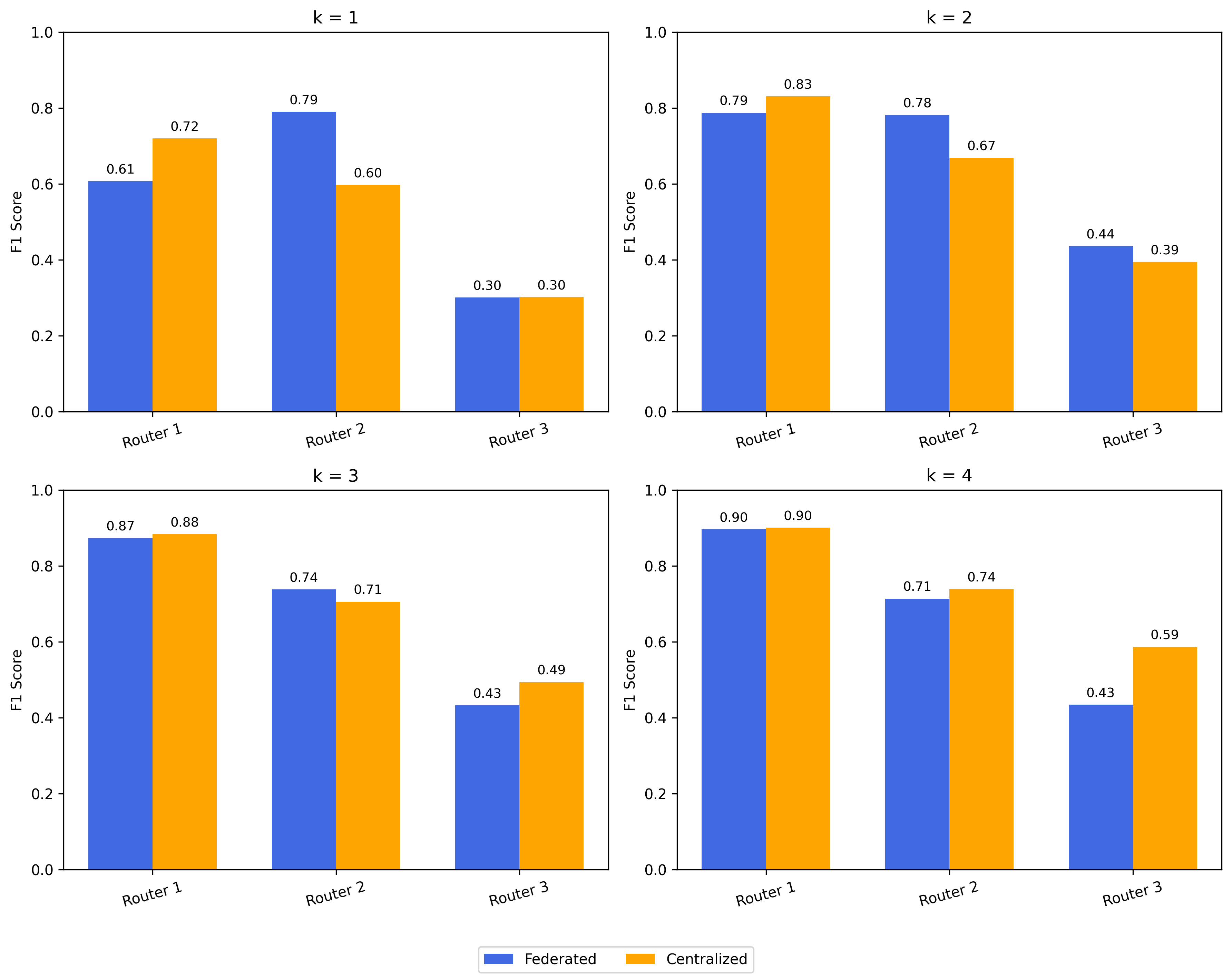}}
         \caption{(a) F1-score for different devices using optimal $k$ values. (b) F1-score for each device for different values of $k \in \{1,2,3,4\}$.}
         \label{fig:LossAndF1}
    \end{figure}

    Table~\ref{tab:combined_metrics} summarizes the evaluation metrics, including accuracy, precision, recall, and F1-score. Additionally, Fig.~\ref{fig:LossAndF1}(b) graphically illustrates the F1-score for the three routers (R1, R2, and R3) across various values of $k$. In this hierarchical setup, the coordinator functions as a global device, while the routers serve as local devices within the network.
    
    For the routers, which represent local devices, variations can be observed across both the federated and centralized models. For Router 1 (R1), both approaches achieve high accuracy, with the best F1-scores at $k = 4$ (0.8963 for the federated and 0.9012 for the centralized model). The results indicate that R1 performs consistently well across all $k$ values, with only minor differences between the two methods.

In the case of Router 2 (R2), both models exhibit similar performance trends, though the centralized model achieves a slightly better balance between precision and recall, especially at higher $k$ values. The optimal performance for R2 in the federated setup occurs at $k = 1$ (F1-score = 0.7895), whereas the centralized model performs best at $k = 4$ (F1-score = 0.7393).

For Router 3 (R3), the results show greater variability across $k$ values. The federated model achieves its highest F1-score at $k = 2$ (0.4358), while the centralized model reaches a significantly higher F1-score of 0.5870 at $k = 4$. The federated approach generally yields higher recall values, suggesting better sensitivity in detecting positive instances, whereas the centralized model attains higher precision, indicating fewer false positives.

Overall, the centralized model demonstrates slightly superior and more stable performance, particularly at $k = 4$, while the federated model performs competitively with router-specific variations.

    \subsection{Optimal Value of $k$}
    The F1-score was instrumental in selecting the most suitable value of $k$ for our system. Through detailed analysis, we identified the optimal $k$ that maximizes the F1-score, ensuring a balanced trade-off between precision and recall. This optimal value was consistently applied across both centralized and federated models to ensure robust and accurate anomaly detection.

    For the federated model, the optimal values of $k$ varied across devices: Router 1 (R1) achieved its best performance at $k=4$, Router 2 (R2) at $k=1$, and Router 3 (R3) at $k=2$. In contrast, for the centralized model, the optimal value of $k$ was consistent across all devices, with all achieving their best performance at $k=4$. These values were selected based on the highest F1-scores obtained for each configuration.

    Fig.~\ref{fig:LossAndF1}(a) shows the F1-scores for optimal k values, which highlights the effectiveness of our federated learning model compared to the centralized counterpart across various devices. The graph illustrates that the federated model achieves competitive F1-scores, demonstrating its capability to detect anomalies efficiently while preserving data privacy and reducing communication overhead.

    \subsection{Communication and Computation Overhead}

    In evaluating federated and centralized learning approaches for IoT anomaly detection, we considered communication and computation overhead as pivotal factors. Communication overhead encompasses data exchanged between devices and the central server or coordinator. Federated learning reduces this by transmitting model updates rather than raw data, enhancing privacy and reducing network congestion. Computation overhead refers to the processing power and time for model training. Centralized learning concentrates computation on a central server, while federated learning distributes load across devices, requiring sufficient local processing capabilities.
    
    Federated learning minimizes communication overhead by transmitting only model updates, conserving bandwidth and reducing data transfer. However, it introduces computation overhead on local devices due to distributed training. Despite this, federated learning leverages collective device processing power, making it efficient for real-time anomaly detection in IoT networks. This approach also enhances data privacy and security, addressing critical IoT concerns. Below, we compare the communication overhead of centralized and federated learning methods over a 5-hour period.
    
    In the centralized method, each edge device transfers data to the routers every second. Over a 5-hour period, each router receives a total of 1.5 MB of data. With 3 routers in the network, the total data transferred is 4.5 MB.
    In the federated learning framework, each router transmits its local model weights (12.6 KB) to the central coordinator every 60 minutes for aggregation. Following the aggregation process, the coordinator distributes the updated global weights of the same size (12.6 KB) back to all routers. Thus, each communication round involves a total bidirectional transfer of 25.2 KB per router. Over a 5-hour training period (five communication rounds), the cumulative data exchanged amounts to approximately 126 KB per router, resulting in a total communication overhead of 378 KB across all three routers. 
    
    It clearly demonstrates that the federated learning method significantly reduces the communication overhead compared to the centralized method. Specifically, federated learning requires only 378 KB of data transfer over 5 hours, whereas the centralized method requires 4.5 MB. This reduction is particularly advantageous in networks with limited bandwidth or where minimizing communication costs is essential. By understanding and managing these overheads, federated learning can serve as an efficient and scalable solution for IoT anomaly detection, contributing to enhanced network intelligence and security.

    \subsection{Complexity Analysis}

    In our anomaly detection system using federated learning on Raspberry Pi 3B+ sensor nodes, the communication overhead involves each edge device sending 12.6 KB of model weights to the coordinator every 60 minutes, and the coordinator sending the updated 12.6 KB global model back to each edge device. Memory complexity for each edge device is $O(m*d)$ for storing local data and $O(d)$ for model weights, while the coordinator requires $O(d + n*d)$ for aggregated weights and the global model, where $m$ is the number of local data points, $d$ is the dimension of each data point, and $n$ is the number of edge devices \cite{rajasegarar2006distributed}. Computationally, training the autoencoder on local data at each edge device involves $O(e*m* d^2)$ operations, with e being the number of epochs, and calculating reconstruction loss is $O(m*d)$. The coordinator's computational complexity for aggregating weights and updating the global model is $O(n*d)$.

\section{Conclusion}
   
In this paper, we evaluated anomaly detection using autoencoder models with both centralized and federated learning approaches. Our results show that the federated approach achieves performance metrics nearly identical to the centralized approach, including accuracy, precision, recall, and F1-score.

Federated learning preserves detection accuracy while enhancing privacy and security. Unlike centralized learning, where data is aggregated at a central point, federated learning processes data locally at individual devices, reducing the risk of data breaches and ensuring sensitive information remains private. We conducted the evaluation using a real network setup with real data and simulated attack scenarios, validating our models for practical IoT environments. In addition to redirection attacks, the proposed method can also detect other attacks, such as Denial of Service (DoS) and man-in-the-middle. Future work will include testing these attacks in real networks.

We also plan to investigate device heterogeneity, evaluate performance in larger networks, and implement adaptive thresholds and incremental learning for continuous evolution. Additionally, advanced federated learning techniques will be explored to improve model performance while reducing communication and computational load. In conclusion, federated learning is a viable and effective approach for anomaly detection in IoT networks, balancing high detection accuracy with privacy and security. Our practical implementation highlights its potential to enhance IoT security.

\bibliographystyle{ieeetr}
\bibliography{ref}

\end{document}